# Enhancing Language Learning through Technology: Introducing a New English-Azerbaijani (Arabic Script) Parallel Corpus

Jalil Nourmohammadi Khiarak[1[0000-0002-1928-9081],*], Ammar Ahmadi[2,†], Taher Akbari Saeed[3,†], Meysam Asgari-Chenaghlu[4,†], Toğrul Atabay[5,†], Mohammad Reza Baghban Karimi[6,†], Ismail Ceferli[7,†], Farzad Hasanvand[8,†], Seyed Mahboub Mousavi[9,†], Morteza Noshad[10,†]

[1,3]Kartal Ol Research Group, Warsaw, Poland
[2]Savalan Igidlari Publishing Company, Ardabil, Iran
[4]Research and Development Ultimate, Berlin, Germany
[5]Institute of Turkic Studies, Marmara University, Istanbul, Turkey
[6]University of Kurdistan, Kurdistan, Iran
[7]SumerNashr Publishing Company, Tabriz, Iran
[8]Department of Mathematics and Statistics, Lorestan University, Lorestan, Iran
[9]Faculty of Engineering, University of Mohaghegh Ardabili, Ardabil, Iran
[10]Computer Science and Engineering, University of Michigan, Michigan, USA
[*]Jalil.kartal@kartalol.com

**Abstract.** This paper introduces a pioneering English-Azerbaijani (Arabic Script) parallel corpus, designed to bridge the technological gap in language learning and machine translation (MT) for under-resourced languages. Consisting of 548,000 parallel sentences and approximately 9 million words per language, this dataset is derived from diverse sources such as news articles and holy texts, aiming to enhance natural language processing (NLP) applications and language education technology. This corpus marks a significant step forward in the realm of linguistic resources, particularly for Turkic languages, which have lagged in the neural machine translation (NMT) revolution. By presenting the first comprehensive case study for the English-Azerbaijani (Arabic Script) language pair, this work underscores the transformative potential of NMT in low-resource contexts. The development and utilization of this corpus not only facilitate the advancement of machine translation systems tailored for specific linguistic needs but also promote inclusive language learning through technology. The findings demonstrate the corpus's effectiveness in training deep learning MT systems and underscore its role as an essential asset for researchers and educators aiming to foster bilingual education and multilingual communication. This research covers the way for future explorations into NMT applications for languages lacking substantial digital resources, thereby enhancing global language education frameworks. The Python package of our code is available at https://pypi.org/project/chevir-kartalol/, and we also have a website accessible at https://translate.kartalol.com/.

[*] Corresponding author
[†] These authors contributed equally to this work

## 1 Introduction

Language learning and machine translation (MT) have undergone transformational changes with the advent of technology, particularly through the development of neural machine translation (NMT) systems [1]. These advancements have significantly improved translation quality between high-resource languages, leading to more effective and accessible language learning tools. However, the benefits of these technological advancements have not been uniformly distributed, with many under-resourced languages remaining neglected in the NMT landscape [2].

The Azerbaijani language, employing both Latin and Arabic scripts [3], exemplifies such under-resourced linguistic scenarios. Despite its rich cultural heritage and significant speaker base, Azerbaijani has seen limited development in terms of digital language resources, particularly for the Arabic script variant [4]. This gap not only hinders the development of effective MT systems but also impacts language learning and educational opportunities for Azerbaijani speakers [5].

In response to this gap, this paper introduces a novel English-Azerbaijani (Arabic Script) parallel corpus, designed to facilitate the development of MT systems and support technological advancements in language learning. The corpus, comprising over 548,000 parallel sentences from diverse sources, represents a significant contribution to the field and aims to address the disparities in language resource availability.

This study also presents the first case study applying NMT to the English-Azerbaijani (Arabic Script) language pair, exploring the efficacy of these systems in low-resource settings and their implications for language education. By leveraging this new dataset, we aim to enhance the understanding of NMT's potential in under-resourced language contexts and contribute to more inclusive language learning technologies.

The remainder of this paper is organized as follows: Section 2 describe motivation of the paper. Section 3 describe used dataset in this paper. Section 4 describes the methodology employed in developing the MT system. Section 5 presents the results of our NMT experiments and discusses their implications for language learning and educational technology. Finally, Section 6 concludes with a summary of our findings and suggestions for future research directions.

## 2 Motivation

The development of a new English-Azerbaijani (Arabic Script) parallel corpus is driven by several critical motivations, reflecting both the specific needs of the Azerbaijani-speaking community and broader challenges in the field of language technology and education.

**Cultural and Linguistic Preservation:** Approximately above 32 million Turks in Iran as shown in Fig. 1, who speak Azerbaijani (or Turkic languages) as their mother tongue, face the risk of linguistic assimilation [6]. The dominance of the Persian language in public and educational spheres poses a threat to the Azerbaijani language and culture, leading to a gradual erosion of linguistic identity. By enhancing the technological resources available for the Azerbaijani language, this research aims to support the preservation and revitalization of this cultural heritage.

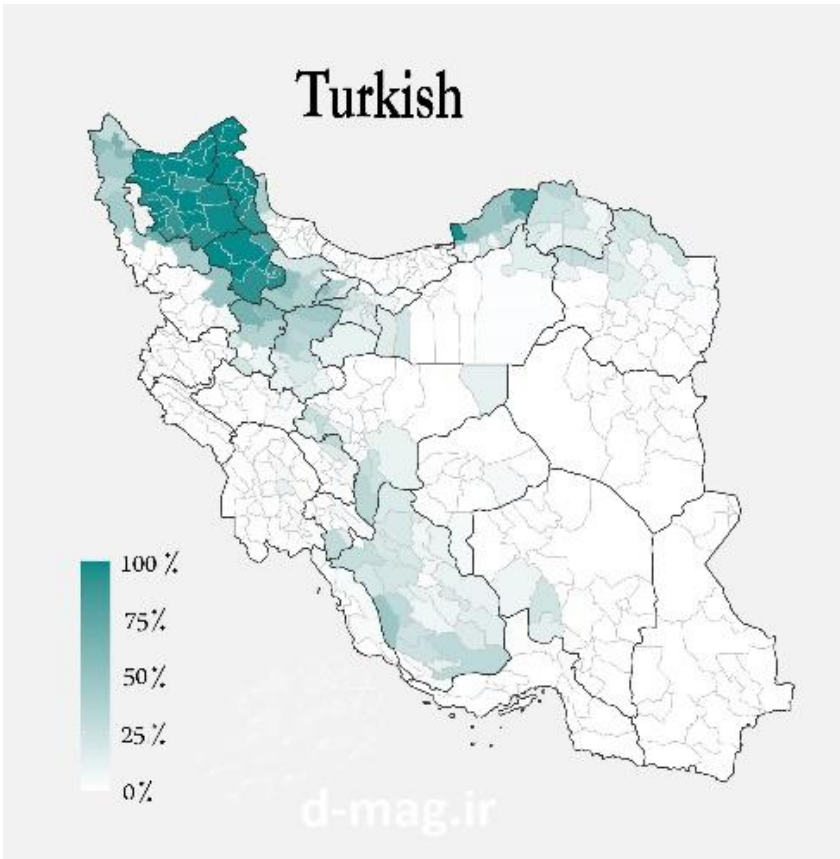

**Fig. 1:** Geographical distribution of Turkic peoples across various provinces and cities in Iran [7]

**Addressing Discrimination:** Linguistic discrimination remains a significant challenge for Azerbaijani speakers, affecting their access to education, employment, and social services [8]. According to Noam Chomsky's theories of language development [9] and universal grammar [10] theory, hindering the development of a child's native language and replacing it with a transitional language, especially during the early years, not only fails to promote cognitive and linguistic (educational) development but also obstructs the growth of the language's original genetic and instinctual basis. Fig. 3 illustrates the distribution of educational attainment among the population in Iran. The development of language technologies, such as machine translation tools, can help mitigate these barriers by improving communication and understanding between different language groups.

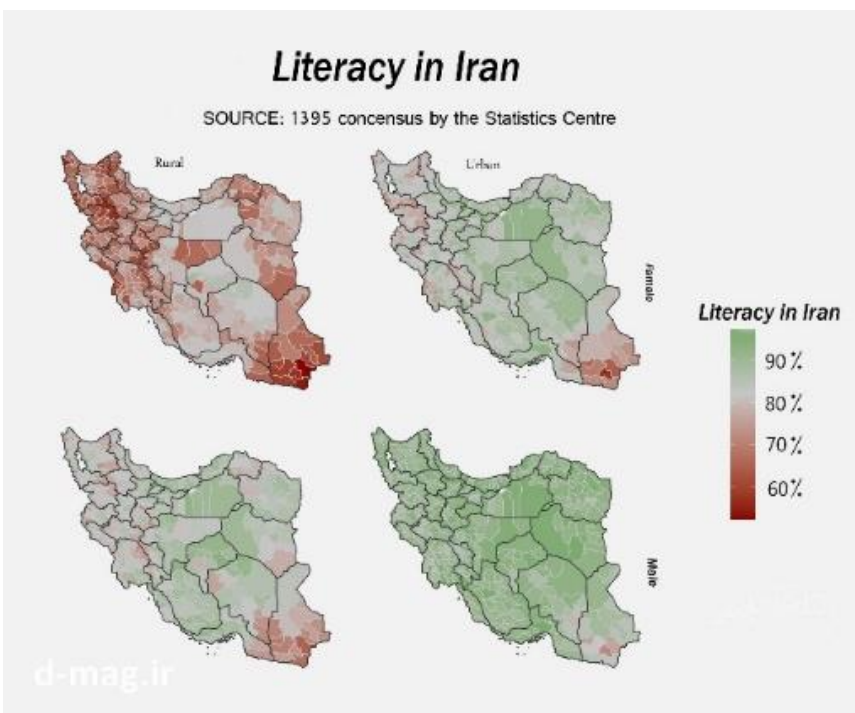

**Fig. 2**: Distribution of Educational Levels within the Iranian Population [7]

**Historical Significance**: The Turkish language is one of the three major languages of the Islamic world (Arabic, Persian, and Turkish) and boasts the largest number of speakers among them. Historically, the Turks have used various alphabets for writing, including the Göktürk alphabet, which dates back 2,500 years. However, following the Islamic conversion of the majority of Turks, the Arabic alphabet was adopted for writing. The earliest known book written in this alphabet (as far as is currently recognized) is the "Dīwān Lughāt al-Turk," penned in 241 AH (856 AD) [11]. In subsequent centuries, efforts were made by Turkish scholars to address the challenges of writing Turkish with the Arabic script. Notably, Al-Zamakhshari (1075–1144 AD) in his book "Muqaddimah al-Adab" attempted to devise rules for easier writing of Turkish using this script, modifying certain Arabic letters [12]. For example, he adapted letters like ط, ض, and ص, which are unique to Arabic, in a way that they could represent strong letters in the Turkish vocabulary. His adaptations were based on a deep understanding of the nuances of the Turkish language. His methodology remained influential until about 50 years ago.

The Azerbaijani language has a rich history, notably becoming the official language of the Azerbaijan People's Government in 1946. Despite its brief tenure as an official language, this period symbolizes the linguistic and cultural aspirations of the Azerbaijani people [13]. Developing digital resources for Azerbaijani not only acknowledges this historical significance but also supports contemporary efforts to recognize and validate the language's status.

In addition, in 1979, with the publication of the prestigious magazine "Varlıq," Dr. Hamid Notghi employed scientific efforts and succeeded in improving the Arabic script for easier writing of the Azerbaijani language [5].

Today, given that the Islamic Republic of Iran government only permits the use of this script and forbids the use of the Latin alphabet, recent years have seen intensified efforts to adapt the Arabic alphabet for writing Turkish texts. Decisions regarding this were made at two conferences attended by Azerbaijani writers and journalists in Tehran in 2004 and 2006. However, further improvements were made following these conferences, and other books emerged from a collective of authors, which today form the basis for the use of this alphabet.

**Research and Technological Advancement:** Beyond these specific motivations, there is a broader imperative to develop resources for under-resourced languages. The field of natural language processing (NLP) has predominantly focused on high-resource languages, leaving a significant gap in technology and research for languages like Azerbaijani [14, 15]. By creating this parallel corpus, this study contributes to a more inclusive and equitable landscape in NLP research, enabling the development of tailored technological solutions that cater to the needs of diverse linguistic communities.

**Educational Impact:** Finally, the availability of robust language resources significantly impacts language learning and educational outcomes. By providing tools that facilitate more accurate and nuanced translations, educators and learners can engage with materials in their native language, enhancing understanding and retention [16]. This research, therefore, holds the potential to transform educational experiences for Azerbaijani speakers, promoting bilingualism and multilingualism as valuable skills in a globalized world.

# 3 Database Description

The TIL Corpus amalgamates an extensive parallel corpus that incorporates the majority of available public datasets across 22 Turkic languages. Presently, the corpus provides data across nearly 400 language pairs, with development and test sets prepared for more than 300 of these directions [1].
Our dataset, was meticulously compiled, incorporating an Azerbaijani to English translation corpus. Initially composed in Azerbaijani utilizing the Latin alphabet, the texts underwent a conversion process to the Arabic script, facilitated by an automated system known as Mirze [17], followed by rigorous verification by a dedicated team from SumerNashr Publishing. This thorough review process ensured each text was scrutinized and validated thrice to uphold accuracy and consistency.
The organizational structure of the database is methodically arranged across multiple files, categorically divided into two principal segments: Azerbaijani (Latin) and English, each comprising approximately 548,000 sentences as described in Table 1. Subsequently, post-conversion of the Azerbaijani content from the Latin to the Arabic script, the resultant files were seamlessly integrated into the existing dataset folder to enhance linguistic diversity and resource completeness. The conversion process from Azerbaijani (Latin) to Azerbaijani (Arabic Script) entailed the extraction of individual words into a document file, which was then meticulously reviewed and amended by the human team to align with standardized orthographic norms [18], ensuring the highest level of linguistic fidelity and scholarly utility.

**Table 1.** Statistical Analysis of Parallel Azerbaijani (Arabic Script) and English Corpus.

| | Azerbaijani (Arabic Script) | English |
|---|---|---|
| Sentences | 548000 | 548000 |
| Words | 9092388 | 11225747 |
| Vocabulary | 171559 | 82701 |

## 3.1 The Orthography for Azerbaijani (Arabic Script)

The Azerbaijani language is regulated according to the orthography rules of 2020, which were prepared in 2020 by the agreement of about 50 writers and experts of this language and after being approved by the Ministry of Culture and Islamic Guidance, were published by Yaznashr, Akhtar, and the journal Farhang-e-Jame'e [18]. Due to the existence of nine vowel letters in the Azerbaijani language, diacritical marks are significantly important in the writing of this language. The most important suggested symbols by Prof. Mohammad Taghi Zehtabi and Prof. Javad Heyat are presented.

Zehtabi, Orthography: (آAa, ئEe, ه,اَ Əə, وْOo, ۆÖö, وُUu and üü, ىْ Iı, ىٌ İi)
Heyat Orthography: (آAa, ئ Ee, ه,اَ Əə, وْOo, ۆÖö, وُUu and üü, يٍ Iı, ى İi)

The 2020 orthography, with the endorsement of both, prefers the symbols suggested by Prof. Zehtabi.

Table 2 illustrates the process and outcomes of converting text from Azerbaijani written in the Latin script to the Arabic script, followed by human corrections to ensure accuracy and adherence to proper language standards. The final row displays the sentence after it has been reviewed and corrected by human experts. These corrections are made to fix any mistakes or inconsistencies introduced during the automatic conversion process. In this case, you can see slight adjustments in the use of letters and diacritics, such as changing "شۆبهه‌لر" to "شُبهه‌لر" and "اۆرگیمدن" to "اۆره‌ییمدن", which more accurately reflect the correct spelling and nuances in the Azerbaijani language when written in the Arabic script.

**Table 2.** Examples of human corrections in dataset translations: a writing perspective.

| Sentence Mode | Sentences |
|---|---|
| Original | « Nə qədər ki , Elvira söhbətdən qaçırdı , ürəyimdən şübhələr çəkilmirdi ki çəkilmirdi » . |
| Converted by Mirze | » نه قدر کی ، ائلویرا صؤحبتدن قاچیردی ، اۆرگیمدن شۆبهه‌لر چکیلمیردی کی چکیلمیردی « . |
| Human correction | » نه قدر کی، ائلویرا صؤحبتدن قاچیردی، اۆره‌ییمدن شُبهه‌لر چکیلمیردی کی چکیلمیردی«. |

Table 3 presents an example of the conversion process of Azerbaijani text from Latin to Arabic script, focusing on the modification of certain terms. The table specifically highlights the adjustments made to replace Russian-influenced words, which are predominantly used in the Republic of Azerbaijan, with their Turkish counterparts more commonly utilized in Iranian Azerbaijani communities. This correction is vital for ensuring the text aligns with the linguistic preferences and cultural nuances of Iranian Azerbaijani speakers, replacing terms borrowed from Russian with their native Turkish equivalents that are familiar to the local population. This third row shows the sentence after human reviewers have corrected it. The corrections reflect more authentic Azerbaijani vocabulary:

"بازار" replaces "مارکئت" (market to bazar), shifting from the Russian-influenced "market" to the native Azerbaijani word "bazar" (market). "یئرآلما" replaces "کارتوف" (kartof to yerəlma), changing from the Russian "kartof" (potatoes) to the Azerbaijani "yerəlma" (earth apple, which means potatoes). "گوجه" replaces "پۆمیدور" (pomidor to goje), substituting the Russian-derived "pomidor" (tomatoes) with the native Azerbaijani "goje" (tomato).

**Table 3.** Examples of human corrections in dataset translations: a writing perspective.

| Sentence Mode | Sentences |
|---|---|
| Original | Biz dünən marketə getdik, kartof, pomidor və səməni aldıq, sonra restorana keçdik. |
| Converted by Mirze | بیز دۆنن مارکئته گئتدیک، کارتوف، پۆمیدور و ثمنی آلدیق، سۆنرا رستورانا کئچدیک. |
| Human correction | بیز دۆنن بازارا گئتدیک، یئرآلما، گوجه و ثمنی آلدیق، سۆنرا رستورانا کئچدیک. |

## 4 The Development of an MT Systems

In this study, we introduce a machine translation framework designed to translate English sentences to Azerbaijani (Arabic script) inspired from [19]. The core of our framework is based on a transformer model.

### 4.1 System Architecture

The Transformer model consists of an encoder-decoder architecture:

- **Encoder:** The encoder processes the input sequence (in English) and generates a semantic representation. It consists of:
  - **Tokenization:** The input text is tokenized into individual words or subwords using a pre-trained tokenizer based on Byte Pair Encoding (BPE) algorithms.
  - **Positional Encoding:** Adds positional information to each token embedding, using sinusoidal functions to encode word positions within the sequence.
  - **Self-Attention Mechanism:** Each layer of the encoder contains a multi-head self-attention mechanism, which computes attention scores for each word concerning every other word. This mechanism consists of 8 heads, each producing attention scores that are concatenated and linearly transformed.
  - **Feed-Forward Neural Networks:** Each layer contains fully connected neural networks, processing the outputs of the self-attention mechanism independently for each position. These networks use two linear layers with ReLU activation in between.
- **Decoder:** The decoder generates the output sequence (in Azerbaijani), leveraging the encoder's representation and incorporating its own layers:
  - **Tokenization:** The Azerbaijani text is tokenized similarly to the input using the same tokenizer.
  - **Target Masking:** To prevent each token from attending to future tokens, a triangular mask is applied.
  - **Self-Attention:** Similar to the encoder, the decoder has a multi-head self-attention mechanism, but it also includes an additional cross-attention mechanism that attends to the encoder's output representation.

- **Feed-Forward Neural Networks:** Like the encoder, each layer contains a fully connected network with two linear layers and ReLU activation.

### 4.2 Training and Inference Procedures

Transformer models are typically trained using parallel corpora, which are pairs of source (English) and target language (Azerbaijani) sentences. The model learns to map input sequences to output sequences by minimizing a loss function such as cross-entropy loss. During training, the model adjusts its parameters (weights) using back propagation and gradient descent algorithms. During training, the model undergoes the following procedures:

- **Loss Function:** The training process minimizes a cross-entropy loss function, comparing the predicted Azerbaijani tokens to the actual tokens.
- **Optimization:** The AdaGrad optimizer is used, adapting the learning rates of each parameter based on their update frequencies. The learning rate is set to 0.01, and other hyper parameters include a dropout rate of 0.1 to prevent over fitting.
- **Source Masking:** During training, a source mask is applied to ignore padding tokens in the input sequence.
- **Training Iterations:** Training proceeds over 50 epochs, or until convergence, using mini-batches of 64 sentence pairs each.

During inference, the model translates new English sentences into Azerbaijani:

- **Token by Token Translation:** The translation process generates tokens one by one, feeding each predicted token back into the model until the end-of-sentence token is produced or a maximum length of 50 tokens is reached.
- **Beam Search:** To improve translation quality, beam search with a beam width of 5 is used, maintaining multiple translation paths and selecting the one with the highest cumulative probability.

This model provides a base model to machine translation from English to Azerbaijani, leveraging the transformer architecture and incorporating detailed mechanisms for tokenization, attention, and optimization. These steps ensure a robust and reproducible translation framework, capable of handling sequence-to-sequence tasks effectively.

## 5 Evaluation

To demonstrate the utility of this corpus in machine translation (MT), experiments were conducted using a test set. The test corpus from the dataset comprises approximately 0.02% of the total sentences.

### 5.1 Hyper Parameters

In comparing the our proposed model with the GPT-4, it's clear that GPT-4 has significantly more parameters (1.76 trillion), indicative of a much larger and potentially more powerful model [20]. The proposed model, with approximately 71 million parameters, is considerably smaller and likely more focused, given its specialized nature for translation tasks. On the other hand, GPT-4's extensive parameter count suggests a model designed for a wide range of general-purpose language understanding and generation tasks, benefiting from deeper and broader learning capabilities. However, the larger size of GPT-4 would also imply greater computational resources for training and inference, which could be a limiting factor for certain applications. In contrast, the proposed model, while smaller, may be more efficient to train and deploy, particularly for specific translation tasks, and could be more accessible for users with limited computational resources.

Table 4 presents the statistics, including the number of sentences, words, and the size of the vocabulary used in the test set. In addition to the aforementioned novelties, we are introducing the first Azerbaijani (Arabic Script) translation gold standard that encompasses all official United Nations languages: Arabic, Chinese, French, Russian, and Spanish. This gold standard contains 4,000 sentences, mirroring the standards provided for other languages in the United Nations v1.0 release [21].

**Table 4.** Azerbaijani (Arabic Script) and English testset corpora statistics.

| Dataset name | Category | Azerbaijani (Arabic Script) | English |
|---|---|---|---|
| KartalOlv.0.testset | Sentences | 2500 | 2500 |
| KartalOlv.0.testset | Words | 10415 | 6550 |
| KartalOlv.0.testset | Vocabulary | 40649 | 49717 |
| UNv1.0.testset | Sentences | 500 | 500 |
| UNv1.0.testset | Words | 13421 | 10845 |
| UNv1.0.testset | Vocabulary | 3956 | 4507 |

Fig. 3 illustrates the evolution of cross-entropy loss over 60 epochs for a Transformer model trained on a machine translation task. The blue line represents the training loss, while the orange line depicts the validation loss. Both losses decrease over time, indicating that the model is learning effectively from the training data. Notably, the training and validation losses converge towards a minimum value of approximately 2.33 and 2.34, respectively, by the end of the 60 epochs. This close convergence, with a negligible gap between training and validation loss, suggests that the model is not overfitting and has good generalization ability on unseen data. The results demonstrate the effectiveness of the Transformer architecture in handling complex language translation tasks while maintaining consistency between training and validation phases.

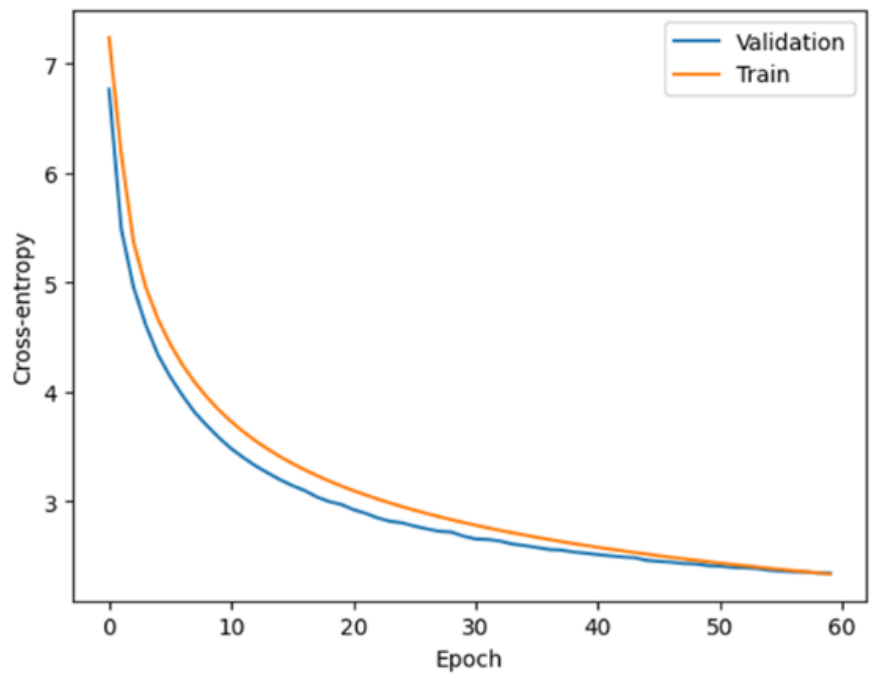

**Fig. 3**: Training and validation loss over epochs for our proposed MT system

### 5.2 Automated Metric Evaluation

In the evaluation segment of this study, we employ three distinct automated metrics: GLEU [23], ChrF [22], and NIST [23], to assess the performance of our machine translation (MT) systems developed using the newly introduced English-Azerbaijani (Arabic Script) parallel corpus. These metrics were chosen due to their varied focus on translation fidelity, fluency, and informative content, providing a comprehensive understanding of system performance across different dimensions (experiments are illustrated in Table 5).

The GLEU metric, an adaptation of the BLEU score specifically tailored by Google, was utilized to gauge the closeness of our translated sentences to a set of reference sentences. Higher GLEU scores indicate a greater alignment with the reference translations, hence denoting superior translation quality. In the context of our test sets, the English-Azerbaijani translations were evaluated against the reference Azerbaijani (Arabic Script) sentences. For instance, within the KartalOlv.0.testset comprising 2,500 sentences, the GLEU evaluation revealed that the translations capture the semantic essence of the English source texts with a notable degree of linguistic precision.

ChrF, focusing on character-level accuracy and fluency, provides an insight into the morphological coherence of the translations. This is particularly significant for Turkic languages, like Azerbaijani, which are agglutinative and where morphological errors can severely distort meaning. Higher ChrF scores indicate a better quality of translation, reflecting not only the correct usage of vocabulary but also the proper construction of words and sentences.

The NIST metric extends beyond mere lexical matching to emphasize the importance of informative content in translations. By assigning greater weight to less frequent n-grams, NIST scores favor translations that effectively convey the meaning of the source text. High NIST scores are indicative of translations that not only are accurate but also retain critical information, an aspect crucial for the educational and informative texts included in our corpus.

**Table 5.** Comparison of evaluation metrics between our model and GPT4 on test sets.

| Test set | Metric | Our model | GPT4 |
|---|---|---|---|
| KartalOlv.0.testset | NIST | **0.25** | 0.145 |
| KartalOlv.0.testset | ChrF | 0.250 | 0.283 |
| KartalOlv.0.testset | GLEU | **0.015** | 0.010 |
| UNv1.0.testset | NIST | 0.11 | 0.151 |
| UNv1.0.testset | ChrF | 0.166 | 0.399 |
| UNv1.0.testset | GLEU | 0.008 | 0.010 |

The evaluation of our English-Azerbaijani (Arabic Script) NMT systems using GLEU, ChrF, and NIST metrics offers a multi-faceted assessment of translation quality. These metrics collectively demonstrate the effectiveness of our corpus in training robust MT systems capable of handling the linguistic diversity inherent in under-resourced languages like Azerbaijani. While GLEU scores highlight the semantic alignment with reference translations, ChrF scores reflect the morphological and syntactic accuracy, and NIST scores underscore the preservation of essential information. In the evaluation of our model against GPT4 using the KartalOlv.0 and UNv1.0 test sets, distinct performance patterns emerge across different metrics. For KartalOlv.0, our model shows a comparable or slightly better performance in NIST and a significantly lower score in ChrF and GLEU metrics compared to GPT4. Conversely, in the UNv1.0 test set, our model underperforms in all metrics, particularly in ChrF, indicating areas for improvement, especially in maintaining content and fluency, as reflected by these metrics. The results underscore the corpus's potential as a pivotal resource for advancing NMT for low-resource language pairs, thereby contributing to the broader objectives of multilingual communication and inclusive language education.

## 6 Conclusion

In conclusion, the English-Azerbaijani (Arabic Script) parallel corpus significantly advances NMT for under-resourced languages, with evaluation metrics indicating varied system performance. While our model demonstrates promising results in certain aspects compared to GPT4, it also highlights areas for improvement. This research underscores the importance of specialized resources for low-resource language pairs and sets a foundation for future advancements in multilingual communication and inclusive language education.

For future research, we plan to extend our work to translate the UN corpus, which comprises approximately 23 million sentences and over 1 million unique Azerbaijani words. This ambitious expansion aims to enhance the depth and breadth of machine translation for Azerbaijani, significantly enriching linguistic resources and improving translation accuracy for this under-represented language.

## Acknowledgements

This work was partially funded by the Kartal Ol research group, and the dataset belongs entirely to this company. The authors extend their gratitude to the Sumer-Nashr and Varliq (Aydin SardarNia) publishing companies for their support and assistance in editing and preparing the dataset. Special thanks also go to Dr. Mehdi Mirza Rasoulzadeh, Head of the Cultural Research Center of Azerbaijan at the University of Mohaghegh Ardabili in Ardabil, Iran, for his guidance and support throughout this project. We also wish to acknowledge Erlik Nowruzi, Senior Deep Learning Scientist at NVIDIA, for his invaluable contributions in revising this paper. Finally, we thank the people of Iranian Azerbaijan for their support and assistance in data collection.